\documentclass{article}

\usepackage[numbers]{natbib}
\usepackage[preprint]{neurips_2022}

\usepackage[dvipsnames]{xcolor}         
\definecolor{linkColor}{rgb}{0.18,0.39,0.62}
\usepackage[utf8]{inputenc} 
\usepackage[T1]{fontenc}    
\usepackage[colorlinks=true,linkcolor=linkColor,citecolor=linkColor,filecolor=linkColor,urlcolor=linkColor]{hyperref}       
\usepackage{url}            
\usepackage{booktabs}       
\usepackage{amsfonts}       
\usepackage{nicefrac}       
\usepackage{microtype}      

\RequirePackage{algorithm}
\RequirePackage{algorithmic}

\usepackage{multirow}
\usepackage{amsmath}
\usepackage{capt-of}
\usepackage{tabularx}
\usepackage{epsfig}
\usepackage{amssymb}
\usepackage{amsfonts}
\usepackage{booktabs}
\usepackage{scalerel}
\usepackage[inline]{enumitem}
\usepackage{listings}
\usepackage{varwidth}
\usepackage[export]{adjustbox}
\usepackage{tikz}
\usetikzlibrary{tikzmark}

\usepackage{stmaryrd}
\usepackage{bbm}
\usepackage{wrapfig}
\usepackage{pifont}

\newcommand{\sptk}[1]{\texttt{[#1]}}

\definecolor{deepblue}{rgb}{0,0,0.5}
\definecolor{officeblue}{RGB}{0,102,204}
\definecolor{deepred}{rgb}{0.6,0,0}
\definecolor{deepgreen}{rgb}{0,0.5,0}
\definecolor{mybrickred}{RGB}{182,50,28}

\definecolor{fillcolor}{RGB}{216,217,252}

\usepackage{amsmath,amsfonts,bm}

\def\eqref#1{equation~\ref{#1}}

\def\1{\bm{1}}

\def\vv{{\bm{v}}}
\def\vw{{\bm{w}}}

\def\mH{{\bm{H}}}

\def\mT{{\bm{T}}}

\def\mV{{\bm{V}}}

\DeclareMathAlphabet{\mathsfit}{\encodingdefault}{\sfdefault}{m}{sl}
\SetMathAlphabet{\mathsfit}{bold}{\encodingdefault}{\sfdefault}{bx}{n}

\newcommand{\R}{\mathbb{R}}

\usepackage{pifont}
\newcommand{\cmark}{{\color{blue}\ding{51}}}%
\newcommand{\xmark}{{\color{red}\ding{55}}}%

\newcommand\our{\textsc{VL-BEiT}}
\newcommand\vlmo{\textsc{VLMo}}
\newcommand\mome{\textsc{MoME}}
\newcommand\beit{\textsc{BEiT}}
\newcommand{\tblidx}[1]{{\small \texttt{[#1]}}}

\title{\our{}: Generative Vision-Language Pretraining}

\author{{Hangbo Bao\thanks{~Equal contribution. $\dagger$ Corresponding author.}, ~~Wenhui Wang\footnotemark[1], ~~Li Dong,
~~Furu Wei$^\dagger$} \\
Microsoft Research \\
\url{https://github.com/microsoft/unilm}
}

\begin{document}

\maketitle

\begin{abstract}
We introduce a vision-language foundation model called \textsc{VL-BEiT}, which is a bidirectional multimodal Transformer learned by generative pretraining. Our minimalist solution conducts masked prediction on both monomodal and multimodal data with a shared Transformer. Specifically, we perform masked vision-language modeling on image-text pairs, masked language modeling on texts, and masked image modeling on images. \textsc{VL-BEiT} is learned from scratch with one unified pretraining task, one shared backbone, and one-stage training. Our method is conceptually simple and empirically effective. Experimental results show that \textsc{VL-BEiT} obtains strong results on various vision-language benchmarks, such as visual question answering, visual reasoning, and image-text retrieval. Moreover, our method learns transferable visual features, achieving competitive performance on image classification, and semantic segmentation.
\end{abstract}

\section{Introduction}
\label{sec:intro}

Generative pretraining has achieved great success in natural language processing~\citep{gpt,bert,unilm,roberta,xlmr,infoxlm} and computer vision~\citep{beit,mae}.
Specifically, BERT~\citep{bert} introduces {masked language modeling}, which learns to recover masked tokens based on the bidirectional contextualized representations encoded by Transformer~\citep{transformer}.
\beit{}~\citep{beit} introduces {masked image modeling} to pretrain vision Transformer~\citep{vit}, which randomly masks image patches and predicts the corresponding visual tokens.

In this work, we explore the mask-then-predict paradigm for multimodal (i.e., vision-language) pretraining.
Our model, namely \our{}, is simple and effective, which is trained from scratch with one unified masked prediction task, one shared Transformer, and one-stage training.
We perform masked prediction on both monomodal (i.e., unpaired images and text) and multimodal data (image-text pairs).
Specifically, the unified objective contains masked language modeling and masked image modeling to learn monomodal representations from large-scale monomodal data, and masked vision-language modeling to aggregate and align visual and linguistic information from multimodal data.
After pretraining, our model can be finetuned on various vision-language and vision tasks.
In addition, we employ mixture-of-modality-experts (\mome{}) Transformer~\citep{vlmo} as the shared backbone network.
Each block of \mome{} Transformer consists of a shared self-attention module across different modalities to align the contents, and a pool of modality experts to capture modality-specific information.
Benefiting from the multimodal pretraining objective and the shared Transformer backbone, \our{} can be used as a image encoder for downstream vision tasks, or finetuned as a dual encoder or fusion encoder for vision-language tasks.

We conduct extensive experiments on vision-language benchmarks including visual question answering, visual reasoning, and image-text retrieval.
Experimental results demonstrate that our model obtains competitive performance across vision-language benchmarks.
We also evaluate our model on vision tasks including image classification and semantic segmentation, achieving strong results.
Ablation studies show that the pretraining tasks and \mome{} Transformer positively contribute to the final performance.

Our contributions are summarized as follows:
\begin{itemize}[leftmargin=1.5em]
\item We introduce a vision-language foundation model named \our{}, which is pretrained by the mask-then-predict task on both multimodal and monomodal data.
\item We propose a simple and effective framework that uses one unified generative pretraining task, one shared bidirectional Transformer, and one-stage training from scratch.
\item Experimental results across various downstream tasks show that our method learns transferable vision-language and visual features.
\end{itemize}

\section{Methods}
\label{sec:methods}

\begin{figure}[t]
\begin{center}
\begin{tabular}{c}
\scalebox{0.96}{
\includegraphics[width=1\textwidth]{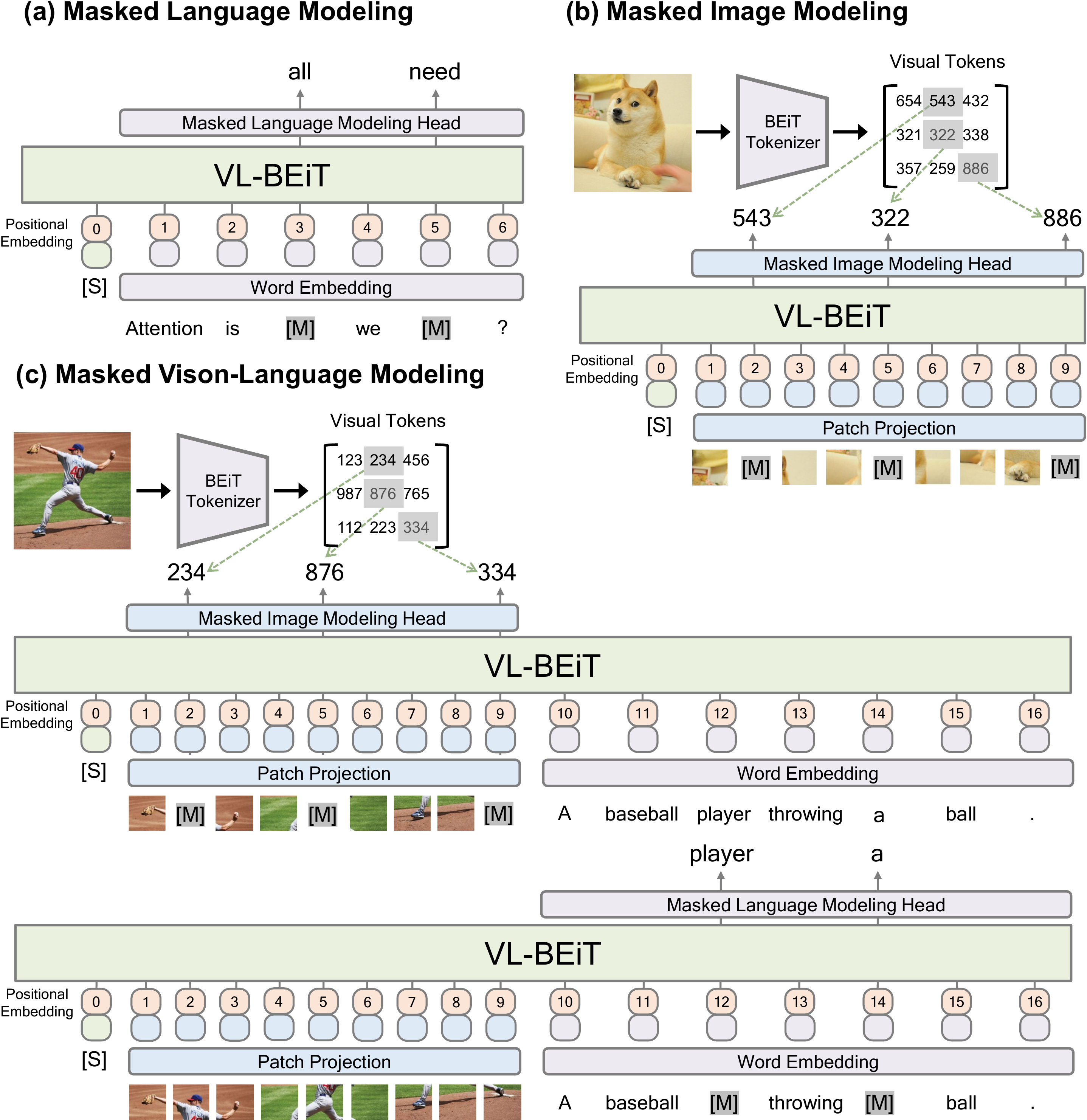}
}
\end{tabular}
\end{center}
\caption{
\our{} is pretrained by masked prediction on both monomodal and multimodal data with a shared Transformer.
}
\label{fig:tasks}
\end{figure}

As illustrated in Figure~\ref{fig:tasks}, \our{} is pretrained by the mask-then-predict task with a shared multimodal Transformer.
We perform masked image modeling on monomodal image data, masked language modeling on monomodal text data, and masked vision-language modeling on multimodal image-text pairs.
After pretraining, the modal can be finetuned as an image encoder, dual encoder, or fusion encoder for various vision and vision-language downstream tasks.

\subsection{Input Representations}

\paragraph{Image Representations}

Following \citep{vit}, we split the image $\vv \in \R^{H \times W \times C}$ into a sequence of patches, so that the image can be encoded by standard Transformer.
The number of patches is $N={HW}/{P^2}$, where $C$ is the number of channels, $(H, W)$ is the image resolution, and $(P, P)$ is the patch resolution.
We then flatten these image patches and obtain patch embeddings ($\{\vv^p_i\}_{i=1}^{N}$) via a linear projection layer.
A learnable special token \sptk{I\_CLS} is prepended to the sequence of patch embeddings.
Finally, we sum image patch embeddings and learnable position embeddings to obtain the final representations $\mH^{v} = [ \vv_{\sptk{I\_CLS}} , \vv_{1} , \dots , \vv_{N} ] + \mV_{pos}$.

\paragraph{Text Representations}

We tokenize the input text and project the tokens to word embeddings ($\{\vw_i\}_{i=1}^{M}$), where $M$ is the length of tokenized text sequence.
Two special tokens, including a start-of-sequence token (\sptk{T\_CLS}) and a special boundary token (\sptk{T\_SEP}), are added to the sequence.
Finally, text representations are obtained via summing the word embeddings and text position embeddings $\mH^{w} = [ \vw_{\sptk{T\_CLS}} , \vw_{1} , \dots , \vw_{M} , \vw_{\sptk{T\_SEP}} ] + \mT_{pos}$.

\paragraph{Image-Text Pair Representations}

Given an image-text pair, we first obtain the image and text input representations as above, respectively. Then we concatenate these vectors to get the image-text pair representations $\mH^{vl} = [\mH^{w} ; \mH^{v}]$.

\subsection{Backbone Network}

We use a shared multimodal Transformer as the backbone network.
Given the image and text representations of monomodal data, and the representations of image-text pairs, we employ a mixture-of-modality-experts (\mome{}) Transformer~\citep{vlmo} to encode different modalities.
Specifically, \mome{} Transformer stacks multiple layers of blocks.
In each block, \mome{} Transformer contains a multi-head self-attention layer and a feed-forward expert layer.
The self-attention module is shared across different modalities.
In contrast, each feed-forward expert layer has a pool of modality-specific experts, which performs as a substitute of the feed-forward network in standard Transformers.
In other words, we use the modality of input token to conduct hard routing over the pool of feed-forward networks.

\mome{} Transformer is flexible to support various downstream tasks by activating different modality-specific experts.
For example, we can use the backbone as monomodal Transformers (i.e., vision or language encoder), multimodal encoders (i.e, with deep fusion), and crossmodal Transformers (i.e., dual encoders).

\subsection{Pretraining Tasks}

\our{} is jointly optimized by masked image modeling on images, masked language modeling on texts, and masked vision-language modeling on image-text pairs.

\paragraph{Masked Language Modeling}

\our{} uses masked language modeling~(MLM) to learn language representations from large-scale text-only data.
Following BERT~\citep{bert}, we randomly mask 15\% tokens of monomodal text data.
Each masked token is replaced by a \texttt{[MASK]} token 80\% of the time, a random token 10\% of the time and kept the original tokens 10\% of the time.
The pretraining objective is to recover the masked tokens from the corrupted input text.

\paragraph{Masked Image Modeling}

In addition to masked language modeling, we employ masked image modeling (MIM) to learn vision representations from large-scale image data.
Following BEiT~\citep{beit}, we apply block-wise masking strategy to mask 40\% of image patches. 
The pretraining objective of MIM is to reconstruct the discrete visual tokens of masked patches.
We use image tokenizer of \beit{}v2~\citep{beitv2} to obtain the discrete tokens as the reconstructed targets.

\paragraph{Masked Vision-Language Modeling}

We introduce masked vision-language modeling (MVLM), which extends masked language modeling and masked image modeling to multimodal data.
The task aims at recovering masked image patches and text tokens based on visual and linguistic clues.
Specifically, we randomly mask text tokens (with 50\% mask ratio) as in MLM, and recover the masked text tokens based on the joint image-text representations.
In addition, we mask image patches as in MIM and predict its corresponding visual tokens based on the image-text pair. The masking strategy is the same as in MIM.
The MVLM task encourages the model to learn alignments between the pairs of image and text.

\section{Experiments}
\label{sec:exps}

We evaluate the pretrained model on vision-language and visual tasks.
We also present ablation studies of pretraining tasks and the backbone architecture.

\subsection{Pretraining Setup}

Our pretraining data consists of monomodal and multimodal data.
For monomodal data, we use ImageNet-22K as the image data, English Wikipedia and BookCorpus~\citep{bookcorpus} as the text data.
The multimodal data combines four datasets of image-text pairs: Conceptual Captions~\citep{gcc}, SBU Captions~\citep{sbu}, COCO~\citep{coco} and Visual Genome~\citep{vg}.
The multimodal data has about $4$M images and $10$M image-text pairs.

Following previous work~\citep{vit,beit,vlmo}, we adopt the same base-size network architecture which consists of $12$-layer Transformer blocks with $768$ hidden size and $12$ attention heads.
We follow the parameter initialization method used in BEiT~\citep{beit}.
The image resolution used for pretraining is $224 \times 224$, and the image patch size is $16 \times 16$.
We mix the data and pretrain the model from scratch with a total batch size of $6,144$ for $480$k steps (i.e., 100 epochs of the image-text pairs).
Each batch contains $2,048$ images, $2,048$ text and $2,048$ image-text pairs.
For the ablation experiments, we train the model for $40$ epochs.
Following \beit{}, we use random resized cropping, horizontal flipping, and color jittering~\citep{coloraug} to perform image augmentation.
We use a SentencePiece tokenizer~\citep{sentencepiece} with $64$k vocab size to tokenize the text data.
Adam~\citep{adam} optimizer with $\beta_1=0.9$, $\beta_2=0.999$ is utilized to optimize the model.
The peak learning rate is 2e-3, with linear warmup over the first $10,000$ steps and cosine learning rate decay.
The weight decay is 0.05.
We disable dropout, and use stochastic depth~\citep{drop_path} with a rate of 0.1.

\subsection{Vision-Language Downstream Tasks}

We conduct vision-language finetuning experiments on the widely used visual question answering~\citep{vqa}, natural language for visual reasoning~\citep{nlvr2} and image-text retrieval~\citep{flickr30k,coco} tasks.
We use $480 \times 480$ image resolution for VQA fine-tuning, and $384 \times 384$ for other tasks.

\paragraph{Visual Question Answering (VQA)} VQA aims to answer questions based on the given image.
Following previous work~\citep{vilt,vlmo}, we use VQA 2.0 dataset~\citep{vqa}, and formulate the task as a classification problem to choose the answer from $3,129$ most frequent answers.
We finetune our model as a fusion encoder to jointly encode the image and question.
The final encoding vector of the \sptk{T\_CLS} token is used as the representation of the image-question pair, and then fed into a classifier layer to predict the label.

\paragraph{Natural Language for Visual Reasoning (NLVR2)}

For visual reasoning task, a text description and a pair of images are given,
the task is to predict whether the description is true about the visual input.
We use NLVR2~\citep{nlvr2} dataset to evaluate the model.
Following OSCAR~\citep{oscar} and VinVL~\citep{vinvl}, we create two image-text pairs based on the triplet input.
Our model is used as a fusion encoder to jointly encode the image and text.
The final vectors of \sptk{T\_CLS} token of the two pairs are concatenated to predict the label.

\paragraph{Image-Text Retrieval}

Depending on the target modality, the task can be divided into two sub-tasks: image-to-text retrieval and text-to-image retrieval.
We use the widely used COCO~\citep{coco} and Flickr30K~\citep{flickr30k} datasets to evaluate the model, and adopt the Karpathy split~\citep{karpathysplit} following common practices.
We employ image-text contrast and image-text matching with hard negative mining objectives as in \vlmo{}~\citep{vlmo} to jointly finetune the model.
During inference, we first use our model as a dual encoder to obtain the top-$k$ candidates, then the model is used as a fusion encoder to rank the candidates based on its image-text matching scores.

\begin{table*}[t]
\centering
\begin{minipage}{2.48in}
\centering
\begin{tabular}{@{}lcccc}
\toprule
\multirow{2}{*}{\bf Model} & \multicolumn{2}{c}{\bf VQA} & \multicolumn{2}{c}{\bf NLVR2} \\
 & test-dev & test-std & dev & test-P \\
\midrule
\multicolumn{5}{l}{\textit{Base-size models pretrained on the same data}} \\
UNITER & 72.70 & 72.91 & 77.18 & 77.85 \\
VILLA & 73.59 & 73.67 & 78.39 & 79.30  \\
UNIMO & 73.79 & 74.02 & - & - \\
ViLT & 71.26 & - & 75.70 & 76.13  \\
ALBEF & 74.54 & 74.70 & 80.24 & 80.50 \\
\vlmo{} & 76.64 & 76.89 & \textbf{82.77} & \textbf{83.34}  \\
\our{} & \textbf{77.53} & \textbf{77.75} & 81.93 & 82.66 \\
\bottomrule
\end{tabular}
\caption{
Finetuning results of base-size models on vision-language classification tasks.
We report vqa-score on VQA test-dev and test-standard split, accuracy for NLVR2 development and public test set (test-P).
}
\label{tbl:results:vl_clssification_tasks}
\end{minipage}
\hfill
\begin{minipage}{2.48in}
\centering
\begin{tabular}{@{}lcccc}
\toprule
\multirow{2}{*}{\bf Model} & \multicolumn{2}{c}{\bf COCO}  & \multicolumn{2}{c}{\bf Flickr30K} \\
 & TR & IR & TR & IR \\
\midrule
\multicolumn{5}{l}{\textit{Fusion encoder}} \\
UNITER & 64.4 & 50.3 & 85.9 & 72.5 \\
VILLA & - & - & 86.6 & 74.7  \\
ViLT & 61.5 & 42.7 & 83.5 & 64.4  \\
\midrule
\multicolumn{5}{l}{\textit{Dual encoder}} \\
\vlmo{} & 74.8 & 57.2 & 92.3 & 79.3  \\
\midrule
\multicolumn{5}{l}{\textit{Dual encoder + Fusion encoder reranking}} \\
ALBEF & 73.1 & 56.8 & 94.3 & 82.8  \\
\our{} & \textbf{79.5} & \textbf{61.5} & \textbf{95.8} & \textbf{83.9} \\
\bottomrule
\end{tabular}
\caption{
Finetuning results of base-size models on image-text retrieval tasks.
We report top-1 recall for image retrieval (IR) and text retrieval (TR).
}
\label{tbl:results:vl_retrieval_tasks}
\end{minipage}
\end{table*}

Table~\ref{tbl:results:vl_clssification_tasks} reports the results on vision-language classification tasks, including VQA and NLVR2.
We compare \our{} with other base-size models pretrained on the same image-text pair data.
\our{} outperforms previous base-size models on VQA and achieves competitive performance on NLVR2.
The unified mask-then-predict pretraining task effectively learns multimodal representations.

Our model also achieves promising performance on image-text retrieval tasks.
As shown in Table~\ref{tbl:results:vl_retrieval_tasks}, we compare with fusion-encoder models, dual-encoder models and the reranking models.
Fusion-encoder models jointly encode all image-text combinations and obtain the similarity scores via the image-text matching objective.
Dual-encoder models encode images and text separately, and compute the similarity scores via a simple interaction layer (i.e., dot product).
The reranking models first obtain topk-$k$ candidates from the dual encoder, and then rank the candidates via the image-text matching scores computed by the fusion encoder.
\our{} outperforms ALBEF, which is also a reranking model, without using image-text contrast/matching during pretraining.

\subsection{Vision Downstream Tasks}

\paragraph{Image Classification}

The task aims to classify the input image to the corresponding category.
We use the ILSVRC-2012 ImageNet dataset~\citep{imagenet}, which consists of $1.3$M images with $1$k classes.
Following \beit{}~\citep{beit}, we perform average pooling over the final vectors, and then feed the resulted vector into a linear classfier layer to predict the label.

\paragraph{Semantic Segmentation} 

The task is to predict the label for each pixel of the input image.
We evaluate our model on the ADE20K dataset~\citep{ade20k}.
The dataset contains $25$K images with $150$
semantic categories.
We use the same task layer as in UperNet~\citep{upernet}.

\begin{table*}[t]
\centering
\begin{tabular}{@{}lcc}
\toprule
\bf Models & \bf ImageNet (acc@1) & \bf ADE20K (mIoU) \\ \midrule
\multicolumn{3}{l}{\textit{Vision Pretraining}} \\
\textsc{ViT}~\citep{vit} & 83.6 & - \\ 
\textsc{BEiT}~\citep{beit} & 85.2 & 52.8 \\ 
\midrule
\multicolumn{3}{l}{\textit{Vision-Language Pretraining}} \\
\our{} & \bf 85.9 & \bf 53.1 \\
\bottomrule
\end{tabular}
\caption{Results of base-size models on image classification (ImageNet-1K) and semantic segmentation (ADE20K).
We report top-$1$ accuracy for ImageNet, and mean Intersection of Union (mIoU) averaged over all
semantic categories for ADE20k.
}
\label{tbl:vision_tasks}
\end{table*}

As shown Table~\ref{tbl:vision_tasks}, we compare \our{} with two base-size vision Transformers on image classification and semantic segmentation.
For \beit{} and \our{}, we perform intermediate finetuning on ImageNet-22k to compare with \textsc{ViT} pretrained on ImageNet-22k.
\our{} outperforms previous state-of-the-art supervised and self-supervised models on ImageNet-1k.
The model also performs competitively on ADE20k.

\subsection{Ablation Studies}

We conduct ablation studies to analyze the contributions of pretraining tasks and \mome{} Transformer used in \our{}.
We evaluate the models on visual reasoning (NLVR2) and image-text retrieval (Flickr30k).

\begin{table*}[t]
\centering
\begin{tabular}{lccc|cccc}
\toprule
& \multicolumn{3}{c}{\textbf{Pretraining Tasks}}&
\multicolumn{2}{c}{\textbf{NLVR2}} & \multicolumn{2}{c}{\textbf{Flickr30k}} \\
& MVLM & MIM & MLM & dev & test-P & TR R@1 & IR R@1 \\
\midrule
\tblidx{1} & \cmark & \xmark  &\xmark & 79.15 & 80.78 & 91.2 & 75.8  \\
\tblidx{2} & \cmark & \cmark  &\xmark & 80.44 & 81.36 & \textbf{92.2} & 77.4 \\
\tblidx{3} & \cmark & \cmark  &\cmark & \textbf{81.10} & \textbf{82.19} & \textbf{92.2} & \textbf{77.9}  \\
\bottomrule
\end{tabular}
\caption{
Ablation studies of pretraining tasks.
}
\label{tbl:ablation:pretraining_tasks}
\end{table*}

\paragraph{Pretraining Task}
Table~\ref{tbl:ablation:pretraining_tasks} presents the results using different pretraining tasks.
Masked image modeling and masked language modeling on monomodal data positively contribute to our method.
In addition, we find that only performing MLM and MIM training on monomodal data gives a relatively low accuracy on NLVR2.
Masked vision-language modeling plays a critical role in our method.

\begin{table*}[t]
\centering
\begin{tabular}{lcccc}
\toprule
\multirow{2}{*}{\bf Architecture} &  \multicolumn{2}{c}{\bf NLVR2} & \multicolumn{2}{c}{\bf Flickr30k} \\
 & dev & test-P & TR R@1 & IR R@1 \\
\midrule
Standard Transformer & 80.77 & 81.42 & 91.7 & 75.8 \\
\mome{} Transformer & \textbf{81.10} & \textbf{82.19} & \textbf{92.2} & \textbf{77.9} \\
\bottomrule
\end{tabular}
\caption{
Ablation study of \mome{} Transformer. 
}
\label{tbl:ablation:mome}
\end{table*}

\paragraph{Backbone Architecture}
We also compare \mome{} Transformer used in our model with standard Transformer.
The results are shown in Table~\ref{tbl:ablation:mome}.
Using \mome{} performs better than standard Transformer on both visual reasoning and image-text retrieval.
Modality experts used in \mome{} effectively capture modality-specific information and improve the model.

\section{Related Work}
\label{sec:related_work}

Vision-language pretraining~\citep{lxmert,vilbert,vl-bert,vinvl,clip,oscar,vilt,albef,simvlm,vlmo,ofa,flamingo,coca} aims to learn multimodal representations from large-scale image-text pairs.
Model architecture and pretraining objectives are critical to the effectiveness of vision-language models.

\paragraph{Model Architectures}
There are two mainstream architectures widely used in previous models: \textit{dual-encoder} and \textit{fusion-encoder} models.
Dual-encoder model~\citep{clip,align} consists of an image encoder and a text encoder. 
It encodes images and text separately, and then employs cosine similarity to model the interaction of image and text vectors.
Dual-encoder models achieve promising results for image-text retrieval tasks with linear time complexity.
However, the simple fusion module is not enough to handle complex vision-language understanding tasks such as visual reasoning.
Fusion-encoder models employ a complex fusion module with cross-modal attention, to jointly encode images and text.
Previous models~\citep{vilbert,vl-bert,oscar,vinvl} use an off-the-shelf object detector like Faster R-CNN~\citep{faster-rcnn} to obtain image region features.
Text features are usually word embeddings or contextual vectors encoded by a text encoder.
These image and text features are then jointly encoded by the fusion module, which usually adopts a multi-layer Transformer network.
Recently, Pixel-BERT~\citep{pixel-bert} and ALBEF~\citep{albef} use CNN/vision Transformer to encode images and remove object detector.
ViLT~\citep{vilt} uses a shared Transformer network to jointly encode image patches and word embeddings.
Fusion-encoder models achieve superior performance on vision-language understanding tasks such as vison reasoning.
But it requires quadratic time complexity for retrieval tasks, which leads to a much slower inference speed than the dual-encoder models.
\vlmo{}~\citep{vlmo} unifies dual-encoder and fusion-encoder models and introduces mixture-of-modality-experts (\mome{}) Transformer to encode various modalities within a shared Transformer block.
In this work, we adopt the \mome{} Transformer as the backbone network given its simplicity and flexibility.
\our{} can also be finetuned as a dual-encoder model or fusion-encoder model.

\paragraph{Pretraining Objectives}
Multiple cross-modal pretraining objectives have been proposed, including image-text contrastive learning~\citep{clip,align}, image-text matching~\citep{lxmert,vilt,albef,vlmo}, masked language modeling~\citep{lxmert,vl-bert,vilt} or prefix language modeling~\citep{simvlm}, masked region classification~\citep{lxmert}, word-patch/region alignment~\citep{uniter,vilt}.
SimVLM~\citep{simvlm} proposes to train the vision-language model using prefix language modeling on image-text pairs and text-only data.
FLAVA~\citep{flava} combines masked image modeling with masked language modeling, image-text contrast and matching based on a fusion-encoder model.
Masked image modeling and masked language modeling are applied on the monomodal encoders.
Masked multimodal modeling, image-text contrast and matching losses are used for the multimodal encoder.
Compared with SimVLM, \our{} introduces richer visual supervision via masked image modeling and masked vision-language modeling.
Different form FLAVA, we use a shared \mome{} Transformer network for different modalities and adopt one-stage training from scratch.

\section{Conclusion}

In this work, we introduce \our{}, a simple and effective approach to pretraining a bidirectional multimodal Transformer encoder for both vision-language and vision tasks.
It solely employs generative pretraining tasks, including masked language modeling on texts, masked image modeling on images, and masked vision-language modeling on image-text pairs. We show that \our{} effectively leverages monomodal data like images and texts as well as multimodal data like image-text pairs. Experimental results show that \our{} gets strong performance on both vision-language and vision tasks.

In the future, we would like to improve \our{} from the following perspectives:
\begin{itemize}[leftmargin=1.5em]
\item We will scale up the model size~\citep{deepnet,xmoe} and data for \our{} training. We would like to explore whether the success of scaling up generative pretraining in NLP can be reproduced for multimodal pretraining under the \our{} framework.
\item Following the research from multilingual language model pretraining~\citep{infoxlm}, we will integrate contrastive objectives like CLIP~\citep{clip} into \our{}, either in pretraining stage by joint learning of generative and contrastive objectives or as an intermediate finetuning task.
\item We are also interested in the zero-shot cross-modality transferability~\citep{zeroshot:modal:transfer} across different modalities like vision and language. 
\end{itemize}

\section*{Acknowledgement}

We would like to acknowledge Zhiliang Peng for the helpful discussions.

\bibliographystyle{plainnat}
\bibliography{univlp}

\end{document}